\documentclass[a4paper, 10pt, conference]{ieeeconf}      
\usepackage{FG2020}
\usepackage{caption}
\usepackage{graphicx, subfig}
\usepackage{amsmath}
\usepackage{multirow}
\usepackage{booktabs}
\usepackage{color}
\usepackage{url}

\FGfinalcopy 

\IEEEoverridecommandlockouts                              
\def\FGPaperID{08} 

\title{\LARGE \bf
Retrieval of Family Members Using Siamese Neural Network
}

\author{\parbox{16cm}{\centering
    {\large Jun Yu, Guochen Xie{$^\dag$}, Mengyan Li and Xinlong Hao}\\
    {harryjun@ustc.edu.cn, \{xiegc,limmy,haoxl\}@mail.ustc.edu.cn}\\
    {\normalsize
    Department of Automation, University of Science and Technology of China, Hefei, China\\}}
    \thanks{{$^\dag$}Corresponding author of this paper. }
}

\begin{document}

\ifFGfinal
\thispagestyle{empty}
\pagestyle{empty}
\else
\author{Anonymous FG2020 submission\\ Paper ID \FGPaperID \\}
\pagestyle{plain}
\fi
\maketitle

\begin{abstract}

Retrieval of family members in the wild aims at finding family members of the given subject in the dataset, which is useful in finding the lost children and analyzing the kinship. However, due to the diversity in age, gender, pose and illumination of the collected data, this task is always challenging. To solve this problem, we propose our solution with deep Siamese neural network. Our solution can be divided into two parts: similarity computation and ranking. In training procedure, the Siamese network firstly takes two candidate images as input and produces two feature vectors. And then, the similarity between the two vectors is computed with several fully connected layers. While in inference procedure, we try another similarity computing method by dropping the followed several fully connected layers and directly computing the cosine similarity of the two feature vectors. After similarity computation, we use the ranking algorithm to merge the similarity scores with the same identity and output the ordered list according to their similarities.
To gain further improvement, we try different combinations of backbones, training methods and similarity computing methods.
Finally, we submit the best combination as our solution and our team(ustc-nelslip) obtains favorable result in the track3 of the RFIW2020 challenge with the first runner-up, which verifies the effectiveness of our method. Our code is available at: {\color{blue} \url{https://github.com/gniknoil/FG2020-kinship}}
\end{abstract}

\section{INTRODUCTION}

Retrieval of family members means finding family members of the given subject, which is a subtask of content based image retrieval(CBIR) task. Different from other subtasks, retrieval of family members takes face images as input to output the similarity order sequence.
As faces are similar in general structure but with delicate differences in illumination, pose and age, this task is always challenging. Recently, many researches\cite{robinson2018visual,wu2018kinship,robinson2017recognizing} have been made in this field and gain much progress. However, the results now are not so satisfying for the inner complexity of the task.

To tackle this issue, we propose our solution. Our solution is composed of two parts: similarity computation and ranking. In similarity computation procedure, we adopt Siamese neural network to compute similarity between two input images. More specifically, two input images are firstly fed into models to produce feature vectors and then the fully connected layers take combinations of features as input to produce a scalar ranging from 0 to 1. In training procedure, the output scalar is viewed as similarity of the image pair and used to compute loss to guide the training. While in general CBIR task, it is typical to use cosine similarity of features as similarity of the image pair. Considering that the two methods are both widely used in many tasks, we compute both for further selection in ranking procedure. In the following chapter, we will refer these two similarities as FC similarity and cosine similarity.
\begin{figure}[t]
    \setlength{\belowcaptionskip}{-0.7cm}
  \centering
  \includegraphics[width=.45\textwidth]{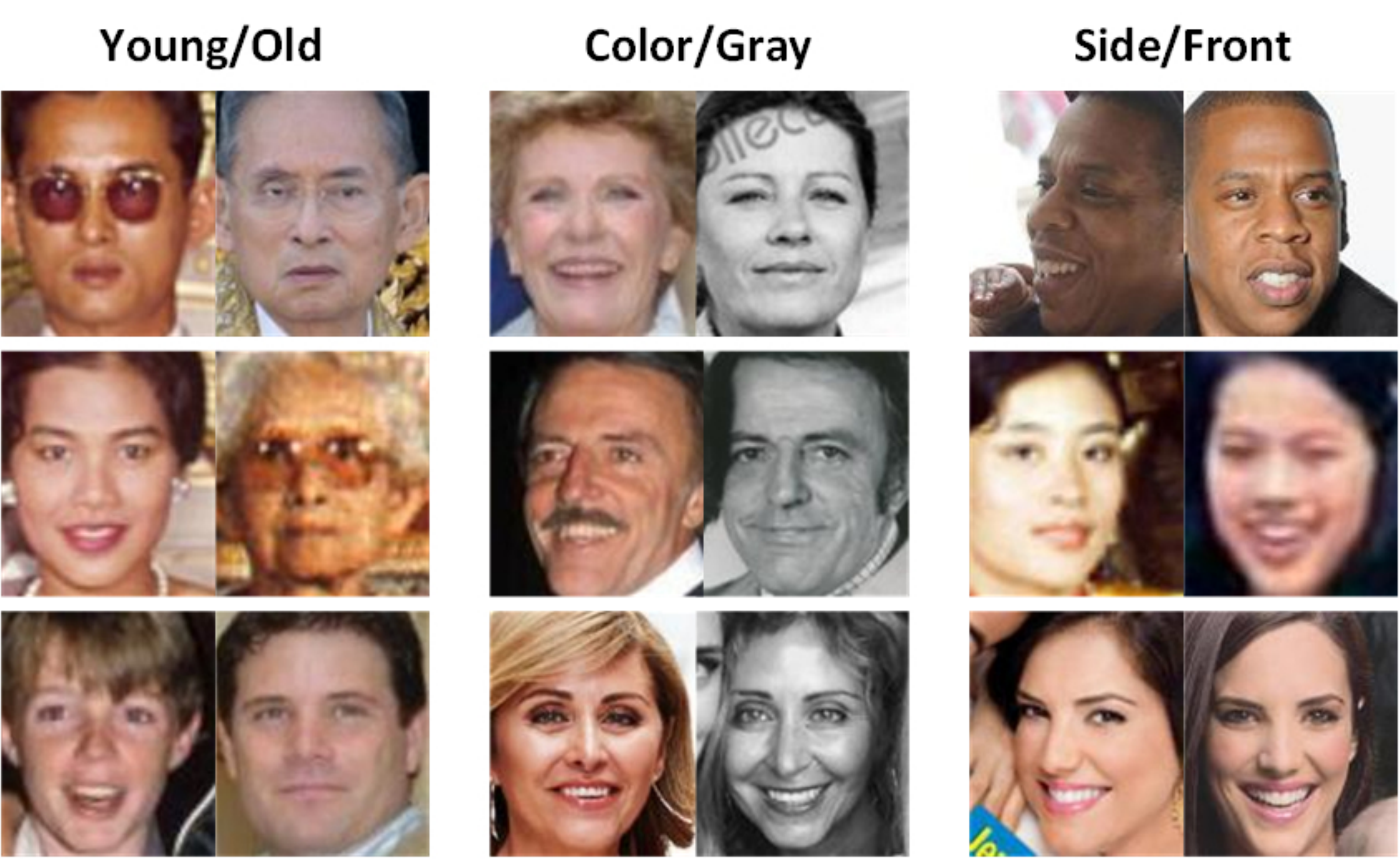}
  \caption{Diversity of the dataset. Each panel indicates the same subject but with different attributes.}
\end{figure}
In ranking procedure, the main purpose is merging similarity scores of the same identity pair and output the ordered sequence. To achieve this goal, we firstly average all the similarities of the same identity pair and then sort them by similarities. In this challenge, we train multiple models with different backbones and loss functions. And for each model, similarity is computed in two ways: cosine similarity or FC similarity. We conduct experiments on the test set and find the best combinations as our solution. Finally, we obtain the first runner-up in the track3 of challenge RFIW2020, which verifies the effectiveness of our method.

To summarize, the main contributions are as follows.
\begin{itemize}
  \item We analyze the inner complexity of the task and composition of provided dataset. As the input images are similar in appearance, we propose our solution based on the Siamese neural network, and achieve favorable results. Different from other methods\cite{lu2014kinship,yan2017kinship}, Siamese network are more capable of modeling delicate and complex relations of image pairs, and thus features extracted with it are more effective and useful.
  \item We try multiple models and training methods for optimizing our solution. Moreover, we build an image retrieval system to find family members of the given subjects, and thus obtaining the first runner-up in the track 3 of the RFIW2020 challenge.
\end{itemize}
\begin{figure*}[thpb]
  \setlength{\belowcaptionskip}{-0.5cm}
  \centering
  \includegraphics[width=.90\textwidth]{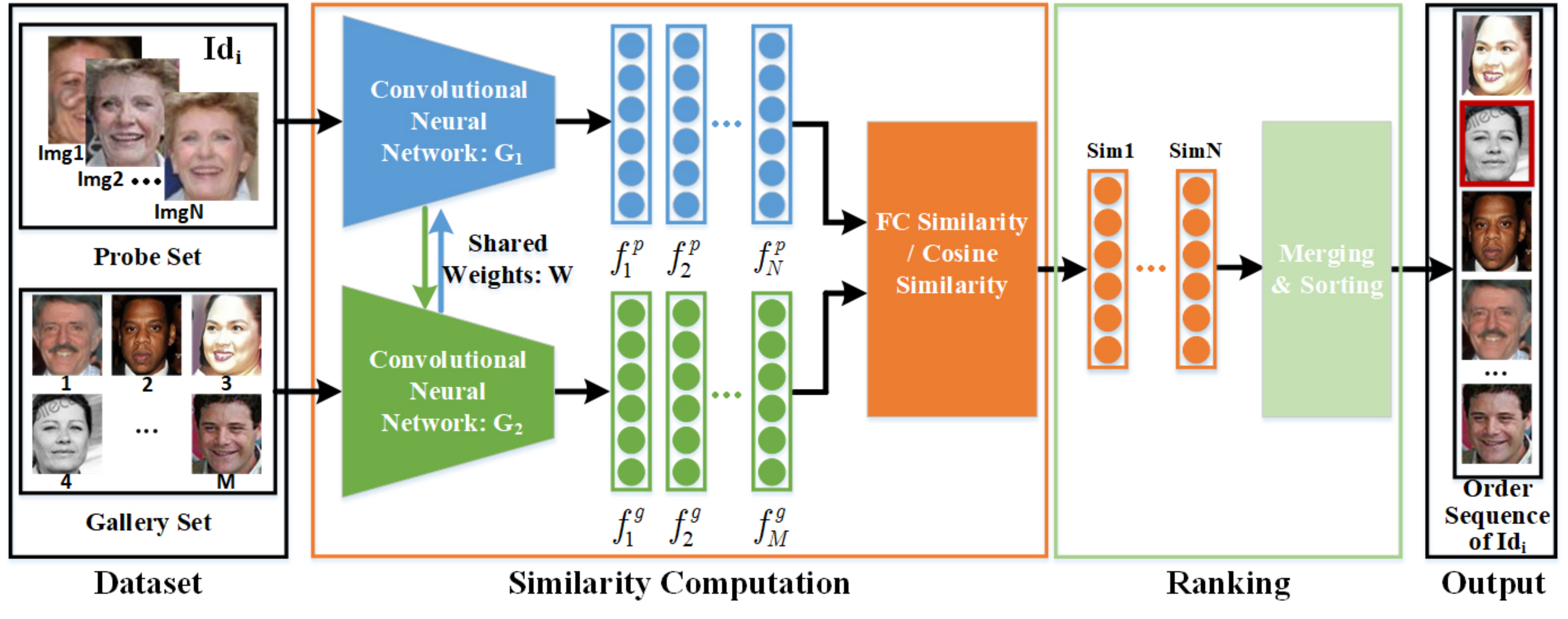}
  \caption{An overview of our solution. Our solution consists of two parts: similarity computation and ranking. In similarity computation procedure, we feed images in the probe and gallery set into the Siamese network($G_1$ and $G_2$) to produce feature vectors($f^{p}$ and $f^{g}$) and then compute similarity(Sim1,$\cdots$,SimN) of inputs with two methods: cosine similarity or FC similarity. Considering each identity typically contains more than one images in probe set(we take Id$_i$ as an example here), we merge all the similarities of the same identity and sort it in the second procedure to output the ordered sequence of retrieval(red box indicates the target image).}
\end{figure*}

\section{RELATED WORK}

\subsection{Deep Learning Based Content Based Image Retrieval}

In recent years, deep learning has been applied to image retrieval to greatly improve results. Oquab et al.\cite{oquab2014learning} propose to use the activations of the fully connected layers as image descriptors. Babenko et al.\cite{babenko2015aggregating} use sum-pooling to extract features of the last convolutional layer and thus propose SPoC method. Similarly, Razavian et al.\cite{sharif2014cnn} replace sum-pooling with max-pooling and proposed MAC method. Most methods above extract features globally and neglect local features. To tackle this problem, Tolias et al. proposed RMAC\cite{tolias2015particular} to integrate both global and local features and thus win great success. All of these methods achieve appealing results in general CBIR tasks. However, as family member retrieval is a special task, in which all the candidates are extremely similar, these methods can't be directly used.

\subsection{Kinship Analysis}

In recent years, kinship analysis has attracted much attention. Roughly, research in this field can be mainly categorized into two classes: kinship verification and family classification. In kinship verification task\cite{wang2017kinship}, Fang et al.\cite{fang2010towards} firstly try to solve kinship verification task with some hand-crafted features.  Following this, Xia et al.\cite{xia2011kinship} find that the great appearance gap will disturb verification. So based on the observation that children's faces look like their parents' at younger ages, subspace transfer learning methods are introduced to bridge the gap between the older faces of parents and younger ones of children. Recently, as deep learning develops quickly and some large datasets such as RFIW\cite{robinson2016families} are made public, many deep learning based methods are proposed. Dahan et al.\cite{dahan2017kin} extract face features with two VGG-Face\cite{parkhi2015deep} models, and then concatenate and feed them to a fully connected network for metric learning. Duan et al.\cite{duan2017advnet} integrated multiple deep face models for kinship verification.

Simultaneously, Family classification also gains much progress as the time elapses. Family classification aims at classifying each subject to one family class. Compared with kinship verification, family classification is more challenging for the large number of family members and great variance. Traditional kinship classification methods\cite{dong2014kinship,fang2013kinship} adopt local feature descriptors(i.e., SIFT\cite{lowe2004distinctive} and HOG\cite{dalal2005histograms}) and focus on capturing the parent-children relationship with complete parents and children data. However, in real-world applications, The data of children is always missing. To tackle this issue, Wu et al.\cite{wu2018kinship} view the children and adults as two separate modalities and thus employ an auxiliary database to transfer knowledge from one modality to the other. In recent years, as deep learning emerges, more and more deeplearning based models(i.e., VGG\cite{parkhi2015deep}, Resnet\cite{he2016deep} and CenterFace\cite{wen2016discriminative}) are applied to family classification tasks and
have shown their capacity in this task. However, there is still much room for improvement.

\section{METHODS}

\subsection{Dataset Description}

In this challenge, the dataset provided is divided by families and identities. Images of the same identity are contained in one folder and all the identities in the same family are included in one previous folder. In the provided dataset, each family has at least two identities and for each identity, there is typically more than one corresponding images. Generally speaking, two major difficulties are introduced with this dataset. The first is that images of the same family vary in pose, image format, age and many other aspects. The visual variance within one family may lead to great difference in feature vectors and do harm to the kinship analysis. Even for images of the same identity, visual difference can also be large(see Fig.1), which further increases the difficulty of this task.
\begin{figure}[thpb]
  \setlength{\belowcaptionskip}{-0.5cm}
  \centering
  \includegraphics[width=.5\textwidth]{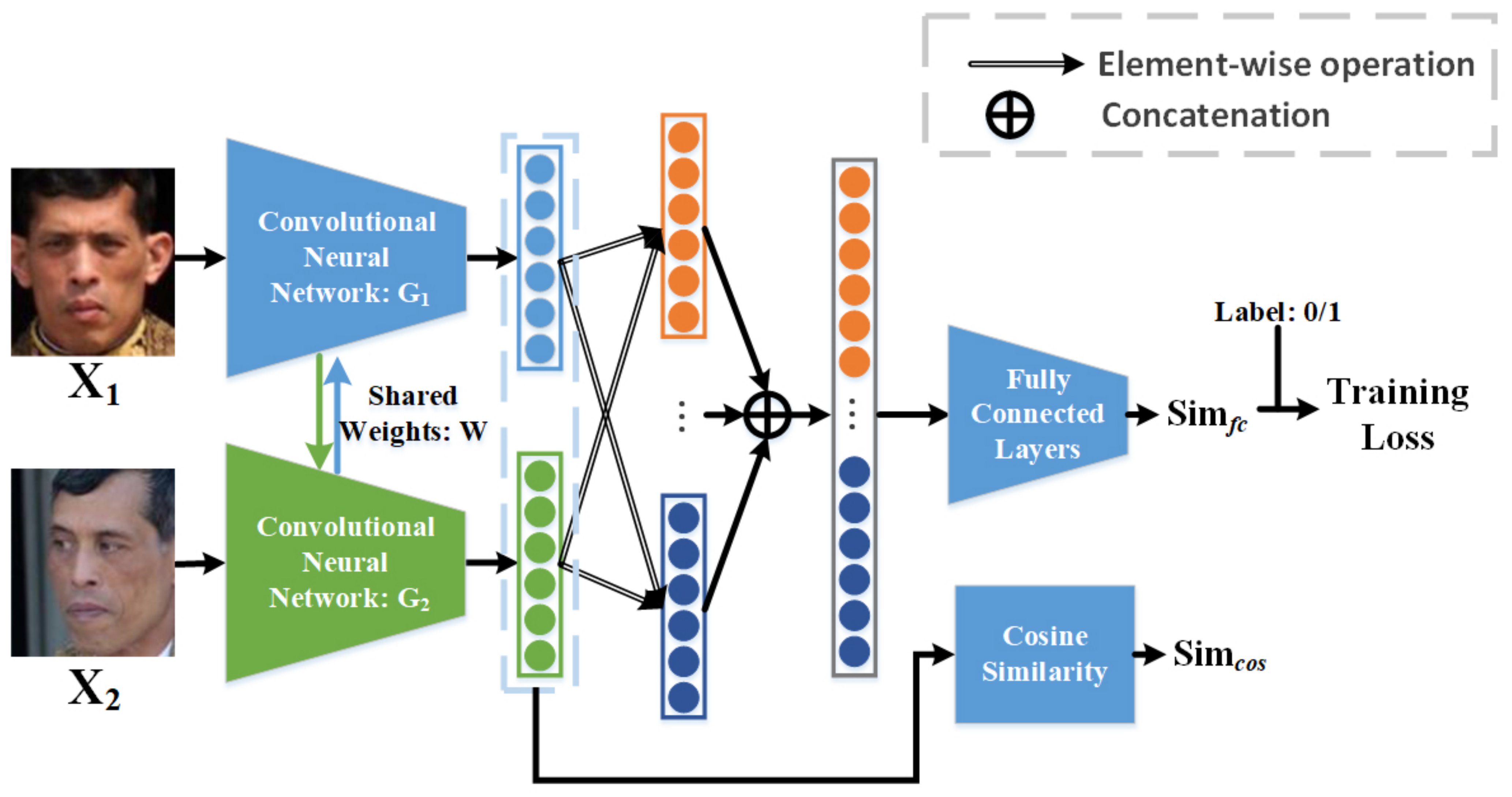}
  \caption{Architecture of the Siamese neural network. Two images are fed into the Siamese neural network to extract features. In training procedure, the features are firstly processed with element-wise operations and then concatenated to form a combined feature.
  Afterwards, the similarity(sim$_{fc}$) is computed with several fully connected layers and compared with labels to guide the training procedure. And in inference procedure, cosine similarity is also adopted by directly compute cosine distance with two feature vectors. The two similarities are both tried for more selection for ranking. }
\end{figure}
Another difficulty is that the dataset contains too many images with the same attributes but of different families. Sometimes, attributes may contribute more to the similarity of two images. Moreover, as some images of the same family are taken under different conditions,
the model is likely to assign large similarity scores for non-family pairs with similar attributes while assigning small ones for family pairs with less common attributes.

Both difficulties may disturb similarity procedure. Thus, we take up building our solution to model the similarity in a more delicate way, and designing ranking method to utilize all the similarity information.

\subsection{Kinship Similarity Computation}

To better model similarities of family members, we adopt a Siamese neural network. As is shown in Fig.3, the Siamese neural network is compose of two branches($G_1$ and $G_2$) with shared weights($W$). After preprocessing, an image pair($X_1$ and $X_2$) is fed into the two branches and then get their corresponding feature maps. And then, feature vectors are obtained by applying average pooling to feature maps. In training procedure, the pairs are sampled previously, while in inference procedure, $X_1$ and $X_2$ are sampled from probe set and gallery set separately. The process can be written as follows:
\begin{equation}
f_1=G_1(X_1),\quad f_2=G_2(X_2)
\end{equation}
where $f_1$ and $f_2$ are extracted features.
A typical siamese neural network computes similarity directly with offline operations of two features. However, this method just models similarity roughly and don't use information of features fully. So, alternatively we adopt network with multiple fully connected layers to model similarity. Moreover, we don't feed the two features into the fully connected layers directly as input but try different combinations of features, i.e., concatenation($\oplus$), addition($+$), etc(see Fig.3). The final output of the whole model is in the scale ranging from 0 to 1. The larger values indicate closer kinship.
This process can be expressed as follows.
\begin{equation}
Sim_{fc}=fc(comb(f_1,f_2))
\end{equation}
where $fc$ means the network used to compute similarities and $comb$ represents all types of feature combination methods. Detailed information about feature combinations is described in the section of experiments.
We use two kinds of loss to guide training procedure: BCE loss and focal loss\cite{lin2017focal}. BCE loss views this problem as a binary classification problem and the output of our model is considered as confidence that they belong to the same family. BCE loss can be written as,
\begin{equation}
\begin{split}
Loss_{BCE}=-label\log(Sim_{fc}) \\
-(1-label)\log(1-Sim_{fc})
\end{split}
\end{equation}
where label is a binary variable indicating whether the pair of image belongs to one family or not.

Focal loss is widely used in face identification and verification tasks, which can be expressed as,
\begin{equation}
Loss_{fl}=
\begin{cases}
-(1-Sim_{fc})^\gamma \log(Sim_{fc})& \text{label=1}\\
-Sim_{fc}^{\gamma} \log(1-Sim_{fc})& \text{label=0}
\end{cases}
\end{equation}
where $\gamma$ is a hyperparameter that can be set manually. In our retrieval process, we firstly extract features of all the images and then compute similarity with cosine distance or via fully connected layers. Cosine similarity is computed as,
\begin{equation}
Sim_{cos}=\frac{f_1\cdot f_2}{\Vert{f_1}\Vert\Vert{f_2}\Vert}
\end{equation}
The two similarities are all evaluated to find the best choice. Detailed analysis is presented in the section of experiments.
\subsection{Ranking Method}

In gallery set, almost all the identities have more than one image. So, for each identity, similarity of multiple images can be utilized to gain better results. In our experiments, we simply average all the similarity scores of the same identity pairs and then sort the similarity sequence to obtain the ordered sequence.

\section{EXPERIMENTS}

\subsection{Implementation Details}
In our experiments, resnet-50\cite{he2016deep} and SENet-50\cite{hu2018squeeze} are adopted as backbones and each is trained with BCE loss and focal loss separately. Thus, we have four types of models in total. For each type of model, we try two different kinds of feature combinations, which is revealed in Table1. Our model is implemented with the Keras\cite{gulli2017deep} framework using Tensorflow backend. A workstation with Intel i7-7700K 4.2G CPU, 64G memory and NVIDIA GTX2080 8G GPU is used for the experiments. Moreover, The model is trained with SGD\cite{bottou2010large} optimizer for 120 epoches and the learning rate is set to 1e-2. Considering the training is conducted with paired data, we randomly select 5000 pairs of family members as positive samples, and then 5000 pairs without any kinship are also selected as negative ones.

The challenge adopts two widely used criterions: mAP and RANK@K. mAP evaluates retrieval performance across recall levels, and RANK@K focuses on if there is any relative images in top-k candidates.
The final rank is based on the average value of RANK@K and mAP(see Table3). And all the scores in our experiments are also calculated in this way. For more detailed description about criterions, we refer you to the website\footnotemark[5] of this challenge.

\renewcommand\arraystretch{1.5}
\begin{table}[tbp]
 \setlength{\abovecaptionskip}{0cm}
 \setlength{\belowcaptionskip}{-0.2cm}
 \caption{Experiment results with FC similarities}
 \center
 \begin{tabular}{cccc}
  \toprule
  Loss & Model Name & FC similarity1\footnotemark[1] & FC similarity2\footnotemark[2] \\
  \midrule
  \multirow{2}{*}{Focal Loss} & Resnet50 & 0.19 & 0.20 \\
  \cline{2-4}
                          & SEnet50 & 0.17 & 0.20 \\
  \cline{1-4}
  \multirow{2}{*}{BCE Loss} & Resnet50 & 0.20 & 0.21 \\
  \cline{2-4}
                          & SEnet50 & 0.18 & 0.20 \\
  \bottomrule
 \end{tabular}
\end{table}
\footnotetext[1]{$(x^2-y^2)\oplus(x-y)^2$}
\footnotetext[2]{$(x^2-y^2)\oplus(x-y)^2\oplus (x\cdot y)$}

\renewcommand\arraystretch{1.5}
\begin{table}[tbp]
\setlength{\abovecaptionskip}{0cm}
\setlength{\belowcaptionskip}{-0.2cm}
 \caption{Experiment results with cosine similarities}
 \center
 \setlength{\tabcolsep}{1mm}{
 \begin{tabular}{cccc}
  \toprule
  Loss & Model Name & cosine similarity1\footnotemark[3] & cosine similarity2\footnotemark[4]\\
  \midrule
  \multirow{2}{*}{Focal Loss} & Resnet50 & 0.19 & 0.21\\
  \cline{2-4}
                          & SENet50 & 0.18 & 0.21\\
  \cline{1-4}
  \multirow{2}{*}{BCE Loss} & Resnet50 & \textbf{0.23} & 0.21\\
  \cline{2-4}
                          & SENet50 & 0.21 & 0.20\\
  \bottomrule
 \end{tabular}}
\end{table}
\footnotetext[3]{Extracting features with max pooling}
\footnotetext[4]{Extracting features with average pooling}
\footnotetext[5]{\color{blue} \url{https://competitions.codalab.org/competitions/22152#learn_the_details-evaluation}}
%
\subsection{Results and Analysis}

In our solution, as similarity computation and ranking are processed in sequence, we can reasonably assume that good retrieval results are more likely originated from accurate models in kinship verification. Thus, based on the observations of the first track of the competition RFIW2020, we choose Resnet50 and SENet50 as our backbones. And both models are trained with focal loss or BCE loss. In our experiments, we adopt two similarity computation method(FC similarity and cosine similarity) and conduct multiple groups of experiments on the provided test set.

The experiments results are shown in Table1 and Table2. In Table1, FC similarity1 and FC similarity2 differ in feature combination method, in which $\oplus$ means concatenation and all the operations(i.e., +, -, $\cdot$) are element-wise operations. In Table2, two feature vectors used in similarity computation are extracted from the output of the last convolutional layer but with different pooling method.

According to the observations, we can find that images of models trained with BCE loss are more effective compared with focal loss. Focal loss is firstly proposed to solve data imbalance problem in training. In our task, the data pairs for training belong to either the same family or different families, which is labeled as positive pairs(1) or negative pairs(0). The organizers provide family labels and identity labels for the whole dataset, and we build our training set by sampling positive pairs and negative pairs from the whole dataset. We sample 5000 pairs for each. Thus, the data imbalance problem is eliminated. In such a condition, we can reasonably think the BCE loss is more suitable compared with focal loss.

Another useful observation is that though the model are trained with FC similarity, but cosine similarity performs better than FC similarity in inference procedure. We review the training and inference procedure and find some reasons. The training set and test set all contain several different kinship relations, and typically, distributions of these relations are not the same. The FC similarities in training set are close to 0 or 1 and can predict labels of data pairs well. But in test set, we find the FC similarities are all small and it is difficult to make any judgement based on these similarities, which means that the several connected layers trained for similarity computation is not so good in generalization and vulnerable to distribution changes. Compared with the FC similarity, cosine similarity directly takes feature vectors for computation and performs more robustly.

\renewcommand\arraystretch{1.2}
\begin{table}[t]
 \setlength{\abovecaptionskip}{0.2cm}
 \setlength{\belowcaptionskip}{-0.2cm}
 \caption{Comparison of retrieval methods on test set of RFIW2020}
 \center
 \setlength{\tabcolsep}{5mm}{
 \begin{tabular}{lccc}
  \toprule
  Team & mAP & Rank@k & Average \\
  \midrule
  vuvko & 0.18 & 0.60 & 0.39(1) \\
  \textbf{ustc-nelslip} & \textbf{0.08} & \textbf{0.38} & \textbf{0.23(2)} \\
  Early & 0.08 & 0.38 & 0.23(2) \\
  DeepBlueAI & 0.06 & 0.32 & 0.19(4) \\
  huunghia160799 & 0.06 & 0.29 & 0.18(5) \\
  danbo3004 & 0.07 & 0.28 & 0.17(6) \\
  \bottomrule
 \end{tabular}}
\end{table}

Among all the combinations, the cosine similarity1 of Resnet50 trained with BCE loss outperforms all others and is submitted as our final solution. Table3 shows some results of competitors. Thanks to the capacity of the Siamese neural network and our reasonable choices, our solution obtains the first runner-up of the track3 with a score of 0.23.
Among all the competitors, our result is just lower than one competitor. And compared with the competitors ranked from the 4th to the 6th. Our solution leads them by a large margin. All of these comparisons verify the effectiveness of our solution.

\section{CONCLUSION}
In this paper, we carefully analyze the requirements and difficulties of track3 in the challenge RFIW2020, and then propose our solution. As human faces are similar in structure and vary in age, pose etc, we adopt the architecture of Siamese neural network to model the delicate difference of human faces. After training, we extract feature vectors with the Siamese network and then compute similarity for ranking. In typical retrieval tasks, there are many different feature extraction methods and similarity computing methods. Thus, we conduct multiple experiments to find the best choice. Finally, our solution obtains the first runner-up in track3 of RFIW2020, which validates effectiveness of our solution.
\bibliographystyle{IEEEtran}
\bibliography{reference}

\end{document}